\pdfoutput=1

\documentclass[11pt]{article}

\usepackage{ACL2023}

\usepackage{times}
\usepackage{latexsym}

\usepackage[T1]{fontenc}

\usepackage[utf8]{inputenc}

\usepackage{microtype}

\usepackage{inconsolata}
\usepackage{graphicx}
\usepackage{subfigure}
\usepackage{booktabs} 

\usepackage{amsfonts}
\usepackage{stfloats}
\usepackage{ulem}

\usepackage{amsmath}

\usepackage{adjustbox}
\usepackage{multirow}
\usepackage{array}
\usepackage{float}
\usepackage{caption}
\usepackage{balance}
\usepackage{flushend}

\usepackage{xurl}

\usepackage{listings}
\usepackage{color}

\definecolor{dkgreen}{rgb}{0,0.6,0}
\definecolor{gray}{rgb}{0.5,0.5,0.5}
\definecolor{mauve}{rgb}{0.58,0,0.82}

\usepackage[group-separator={,}]{siunitx}

\title{Transferring General Multimodal Pretrained Models to Text Recognition}

\author{Junyang Lin, Xuancheng Ren, Yichang Zhang, Gao Liu, Peng Wang, An Yang, Chang Zhou \\ 
DAMO Academy, Alibaba Group \\ 
\texttt{junyang.ljy@alibaba-inc.com}
}

\begin{document}
\maketitle
\begin{abstract}

This paper proposes a new method, OFA-OCR, to transfer multimodal pretrained models to text recognition. 
Specifically, we recast text recognition as image captioning and directly transfer a unified vision-language pretrained model to the end task. 
Without pretraining on large-scale annotated or synthetic text recognition data, OFA-OCR outperforms the baselines and achieves state-of-the-art performance in the Chinese text recognition benchmark. 
Additionally, we construct an OCR pipeline with OFA-OCR, and we demonstrate that it can achieve competitive performance with the product-level API.  
The code\footnote{\url{https://github.com/OFA-Sys/OFA}} and demo\footnote{\url{https://modelscope.cn/studios/damo/ofa_ocr_pipeline/summary}} are publicly available.
\end{abstract}

\section{Introduction}

Optical character recognition (OCR) plays an important role in the real-world applications. It helps users or developers extract text contents from different types of images, including photos, scanned documents, etc. 
In practice, building a tool for OCR needs a pipeline consisting of a text localization module and a text recognition module. 

In this work, we focus on improving the accuracy of text recognition. 
Text recognition has often been regarded as a key challenge owing to the room for improvements in recognition accuracy. 
In the deep learning era, the classical methods are mostly based on CNN and RNN, which are responsible for visual feature extraction and sequence modeling, respectively~\citep{crnn, aster, moran}. 
Recently, with the rise of Transformer~\citep{transformer}, researchers applied the Transformer encoder-decoder framework to text recognition and achieved outperforming results over the baselines~\citep{trocr,maskocr}. 
However, most methods are based on large-scale pretraining on human-annotated or synthetic OCR data. 
It is hard for other researchers to collect or create such data for reproduction. 
Furthermore, the methods often include complex model or objective designs, like DETR-like decoder~\citep{detr}, CTC loss~\citep{ctc}, etc. 
These components also might hinder reproduction as they increase the difficulty in training. 
Therefore, we naturally raise a question: \textit{Is there any way to achieve high recognition accuracy without complex designs on data and model?}

\begin{figure}[t] 
    \centering
    \includegraphics[width=1.0\linewidth]{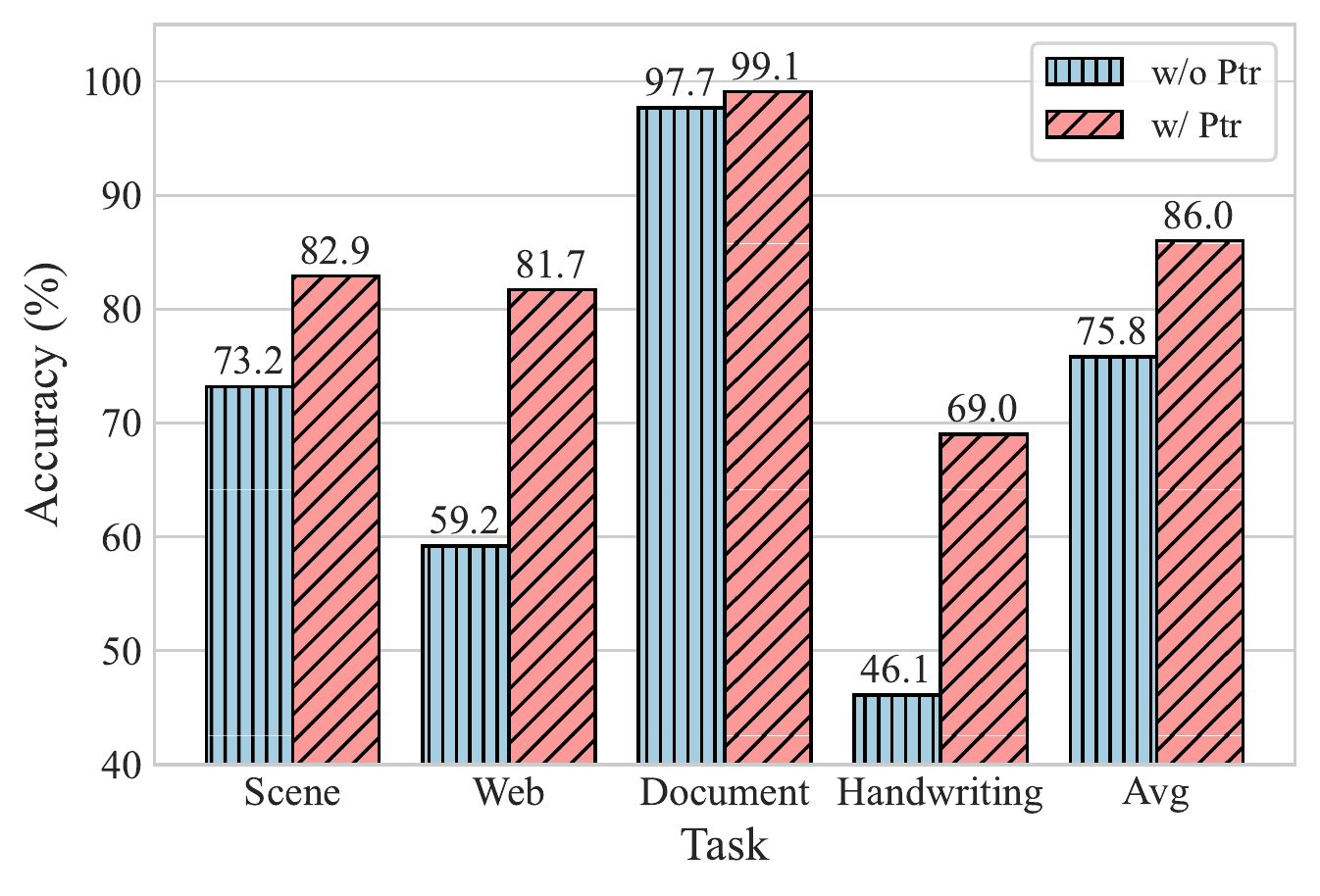}
    \caption{\textbf{An comparison between the performance with or without general vision-language pretraining.} On 4 subtasks of text recognition, OFA-OCR with general-domain vision-language pretraining significantly outperforms the baseline without one. }
    \label{fig:woptr}
\end{figure}

Inspired by the recent progress in multimodal pretraining, we argue that the transfer of a unified multimodal pretrained model is a possible solution. 
Multimodal pretraining has proved significant to the performance of downstream tasks, and thanks to the rise of unified multimodal pretrained models, they can perform both cross-modal understanding and generation and achieve state-of-the-art performance~\citep{ofa, beit3, unified-io}. 
We therefore propose to transfer the unified multimodal pretrained model by finetuning the pretrained model on the text recognition datasets with the task of image captioning, which is essentially a simple sequence-to-sequence learning task with maximum likelihood estimation for optimization. 

To support the effectiveness of the proposed method, we have conducted extensive experiments on the Chinese text recognition benchmark~\citep{benchmarking_chinese} covering multiple scenarios, including scene, web, document, and handwriting. 
Specifically, we finetune the open-source Chinese multimodal pretrained model OFA~\citep{ofa} on text recognition, and we name the model OFA-OCR. 
Figure~\ref{fig:woptr} demonstrates the results of methods with or without general-domain pretraining. 
It shows that multimodal pretraining on general-domain vision-language data can effectively boost downstream performance in text recognition. 
To achieve the best performance, we apply the multitask + single-task finetuning to OFA-OCR, and it outperforms the previous state-of-the-art methods on the benchmark. 
Furthermore, through the ablation studies, we demonstrate the effectiveness of our method designs, including multitask + single-task finetuning, data augmentation, etc. Furthermore, to enable deployment for real-world applications, we construct a pipeline with both OFA-OCR and a simple text localization module. 
We find that this simple pipeline can provide high-quality OCR performance, competitive with a product-level API. 

\section{Method}
\label{sec:method}



\subsection{Pretraining}
To leverage the capability of the multimodal pretrained model for image captioning, we employ the unified multimodal pretrained model architecture. 
Specifically, we implement our models on OFA~\citep{ofa}, an open-source state-of-the-art unified multimodal pretrained model with the release of Chinese models. 

The model is mainly based on the Transformer encoder-decoder framework~\citep{transformer}. 
To make information from different modalities adaptable to the Transformer, there are adaptors for images and texts, which are visual backbones, e.g., ResNet~\citep{resnet}, ViT~\citep{vit}, etc., and word embeddings, respectively. 
The information from modalities is encoded as discrete tokens so that the decoder can perform their generation. 

For Chinese multimodal pretraining, OFA-Chinese was pretrained on a large-scale dataset, which consists of LAION-5B~\citep{laion5b}, Wukong dataset, as well as translated datasets from MSCOCO~\citep{coco_cap}, Visual Genome~\citep{vg}, VQA~\citep{vqav2}, RefCOCO~\citep{refcoco}, etc. 

Note that this work is different from previous pretraining-related methods, which pretrain the model on large-scale human-annotated or synthetic data. 
We show that through pretraining on general-domain data, the model can obtain the potential of text recognition by finetuning on small datasets.

\subsection{Finetuning with Image Captioning}
It is natural to recast text recognition as image captioning, as text recognition also requires the model to generate a piece of text based on the input image. 
It is equivalent to finetuning on different image captioning datasets, where the target refers to the text on the image. 
We finetune the model with maximum likelihood estimation for optimization. 

Furthermore, to better alleviate the discrepancy between upstream and downstream data, we apply a transformation to the input images to make them square, e.g., a resolution of $480 \times 480$. 
Specifically, we first resize the image to a longer edge of the specified resolution while keeping the original height-width ratio of the image, and we make the image square by padding on all sides with the edge value. The lengths for the directions are random, and thus this method can play as data augmentation in this context. 
We demonstrate the pseudo code in Sec.~\ref{implementation_details}. 

For better performance in the downstream tasks, we often use a larger resolution in the finetuning stage, and thus we encounter issues with the positional embedding. 
In our practice, we still use the same one from pretraining but apply interpolation to adapt to images of a larger resolution.

\subsection{Multitask Finetuning}
There are multiple subtasks in text recognition, concerning different scenarios, e.g., scene, document, etc. 
Our experiments are implemented on the Chinese text recognition benchmark consisting of $4$ subtasks. 
In our practice, we implement multitask finetuning and single-task finetuning for comparison. 
Specifically, as the data of all subtasks are organized with the same format, we directly build a mixture of datasets for multitask finetuning. 
We find that directly applying multitask finetuning can help OFA-OCR achieve outstanding performance on all datasets. 
To further boost its performance, we additionally apply single-task finetuning after multitask finetuning, and we find that this pushes its performance to the new state-of-the-art.

\section{Experiments}

\begin{table*}[t]
\center
\small
\vskip 0.15in
\begin{adjustbox}{max width=1.\textwidth}
\begin{tabular}{@{\extracolsep{\fill}}lccccc}
\toprule
  Metrics & Scene & Web & Document & Handwriting & Average
  \\
\midrule
  CRNN~\citep{crnn} & 53.4 & 54.5 & 97.5 & 46.4 & 67.0 \\
  ASTER~\citep{aster} & 54.5 & 52.3 & 93.1 & 38.9 & 64.7 \\
  MORAN~\citep{moran} & 51.8 & 49.9 & 95.8 & 39.7 & 64.3 \\
  SAR~\citep{sar} & 62.5 & 54.3 & 93.8 & 31.4 & 67.3 \\
  TransOCR~\citep{transocr} & 63.3 & 62.3 & 96.9 & 53.4 & 72.8 \\
  $\text{MaskOCR}\rm_{ViT\text{-}B}$ & 73.9 & 74.8 & 99.3 & 63.7 & 80.8 \\
  $\text{MaskOCR}\rm_{ViT\text{-}L}$ & 76.2 & 76.8 & 99.4 & 67.9 & 82.6 \\
\midrule
  $\text{OFA-OCR}\rm_{Base}$ & 82.9 & 	81.7 &	99.1 &	69.0 & 86.0 \\
  $\text{OFA-OCR}\rm_{Large}$ & 83.7 & 	82.6 &	99.2 &	67.7 & 86.3 \\

\bottomrule
\end{tabular}
\end{adjustbox}
\caption{\textbf{Experimental results on the Chinese text recognition benchmark. }Results show that the base-size OFA-OCR model can outperform the previous state-of-the-art, and the large-size model achieves the best performance on average. }
\label{tb:results}
\end{table*}


\subsection{Datasets and Metrics}
We implement OFA-OCR on the Chinese text recognition benchmark~\citep{benchmarking_chinese}. This benchmark consists of multiple subtasks of text recognition, which are text recognition in different scenarios, including scene, web, document, and handwriting. 
The details of the datasets are provided in Sec.~\ref{datasets}. 
The evaluation metric includes accuracy, which refers to the ratio of exact match.

\subsection{Experimental Results}
The experimental results are demonstrated in Table~\ref{tb:results}. 
We compare our method with baseline models of OCR, including the previous state-of-the-art MaskOCR~\citep{maskocr}. 
It can be found that with no regard to the scale of models, the base-size OFA-OCR, which is finetuned from the pretrained Chinese $\text{OFA}_\text{Base}$, can outperform both the base-size and large-size MaskOCR models. 
Specifically, it shows the advantages of $9.0$, $6.9$, and $5.3$ absolute improvements in the scenarios of scene, web, and handwriting. 
On average, the base-size OFA-OCR outperforms the base-size MaksOCR by $5.2$ and the large-size MaskOCR by $3.4$. 
Scaling up the model size can consistently bring steady improvement in the downstream performance. 
On average, $\text{OFA}_\text{Large}$reaches the best results of $86.3$. 

Specifically, we find that the advantage in the scene dataset is the largest among the tasks. 
This may be attributed to the pretraining on general-domain data, where there are images of street views, and some of them might contain texts. 
Similarly, the pretraining dataset consists of web images that resemble those in the web dataset, and thus the gaps between OFA-OCR and the previous methods are large. 
However, text recognition for documents should be a simpler task as the texts are more regular in fonts and there is often much less noise in the background. Thus, even the conventional method like CRNN can achieve a high accuracy.

\subsection{Ablation Study of Training Strategies}
\begin{figure}[t] 
    \centering
    \includegraphics[width=1.0\linewidth]{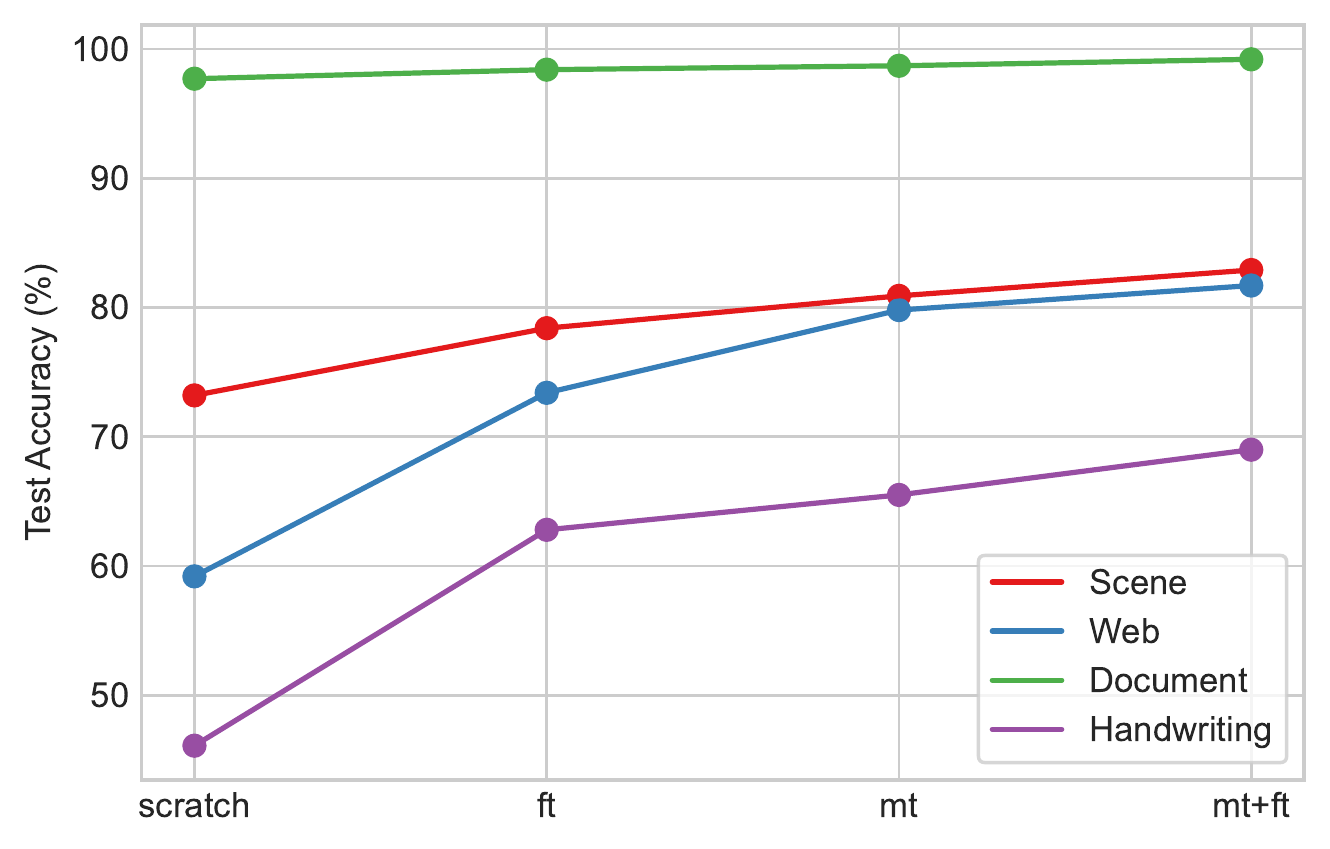}
    \caption{\textbf{Performance of OFA-OCR in different setups. }We validate the model performance on the $4$ datasets in the setups of training from scratch (scratch), single-task finetuning (ft), multitask-finetuning (mt), and multitask + single-task finetuning (mt+ft). We observe consistent performance growth with the addition of the pretrained weight initialization and multitask finetuning. }
    \label{fig:curve}
\end{figure}

To check how the multitask learning influences the final performance, we conduct an ablation study to evaluate its effects. Specifically, the experiments are conducted with the base-size OFA-OCR. 
We provide experiments in $4$ setups, which are training from scratch (scratch), single-task finetuning (ft), multitask-finetuning (mt), and multitask + single-task finetuning (mt+ft), respectively. 
Experimental results are shown in Figure~\ref{fig:curve}. 
It can be found that on average, the addition of the initialization of the pretrained OFA model significantly boosts the performance on the datasets. 
Surprisingly, multitask finetuning alone can outperform single-task finetuning on all $4$ tasks, and the advantage in the web dataset is the most obvious. 
We assume that this is attributed to the small amount of supervised training data for downstream transfer. 
A mixture of datasets of related subtasks can encourage performance on all subtasks. 
Furthermore, the combination of multitask finetuning and single-task finetuning is the best solution owing to its outstanding performance, while multitask finetuning on the mixture of datasets is the most cost-efficient. 

\subsection{Ablation Study of Data Augmentation}
The preprocessing of images for this task can play as data augmentation. 
To validate its effects, we use a simple resizing to the specified resolution as a baseline. 
We also implement experiments on the $4$ datasets, and for simplicity we implement the experiments in the setup of single-task finetuning on the base-size models. 
Results are demonstrated in Table~\ref{tb:aug_exp}. We use ``Aug.'' to indicate the preprocessing method mentioned in Sec.~\ref{sec:method}. The results indicate that the introduced technique for data preprocessing can effectively boost the performance. 

\subsection{Deployment}
To construct an OCR system applicable in real-world scenarios, a strong text recognition model is not sufficient, and we need to build a pipeline with both the text detection and text recognition module. 
While the former one is not the focus of this research, we directly use a light-weight model from EasyOCR\footnote{\url{https://github.com/JaidedAI/EasyOCR}} for detection. After detecting all the bounding boxes which possibly contain texts, we crop them with boxes to create a batch of new images. 
The final step is to process the images with OFA-OCR for the generation of text recognition results. 
Through our case study, we find that the simple OCR pipeline based on OFA-OCR can achieve competitive performance with the product-level API. Examples are demonstrated in Sec.~\ref{case_study}.


\begin{table}[t]
  \center
  \small
  \vskip 0.15in
  \begin{adjustbox}{max width=1.\textwidth}
  \begin{tabular}{@{\extracolsep{\fill}}lcc}
  \toprule
    Method & w/o Aug. & w/ Aug.
    \\
  \midrule
    Scene & 77.0 & 78.4 \\
    Web & 72.3 & 73.4 \\
    Document & 98.2 & 98.4 \\
    Web & 60.4 & 62.8 \\
  \midrule
    Avg & 81.0 & 82.1 \\
  \bottomrule
  \end{tabular}
  \end{adjustbox}
  \caption{\textbf{Performance comparison with or without data augmentation for images. }The experiments are conducted in the setup of single-task finetuning on the base-size model. }
  \label{tb:aug_exp}
\end{table}

\section{Related Work}
We focus on the review of text recognition methods and multimodal pretraining. 
Conventional methods based on CNN and RNN have demonstrated great effectiveness~\citep{crnn, moran, aster, semantic_reasoning, sar, abinet}. 
Recent methods have turned to the use of Transformer and achieved improved performance~\citep{vit_ocr, trocr, conclr, maskocr}. 
However, before this work, we have not witnessed the direct transfer of general-domain vision-language pretrained models to text recognition.
Vision-language pretraining has proved a success as it has leveled up the model performance on a series of downstream tasks~\citep{uniter, vilbert, clip, simvlm}, and the unified models capable of both understanding and generation have become popular and achieved the best performance~\citep{ofa, beit3}. 
Yet, there are only a few unified multimodal pretrained models in Chinese~\citep{m6, ofa}.

\section{Conclusion}

In this work, we propose a simple method called \textbf{OFA-OCR}, which leverages the unified multimodal pretrained model and transfers it to text recognition by image captioning. 
To be more specific, we utilize the Chinese multimodal pretrained model OFA without pretraining on OCR data and transfer it to text recognition with multitask + single-task finetuning.  
Through extensive experiments, we demonstrate that OFA-OCR can achieve state-of-the-art performance on the Chinese text recognition benchmark. 
Additionally, we build a pipeline of OCR by integrating an existent simple text detection module and OFA-OCR. 
The deployed pipeline achieves competitive performance in comparison with a product-level API. 
We hope that this research sheds light on the application of general-domain multimodal pretraining, and also helps OCR practitioners.

\section*{Limitations}
This section discusses the limitations of this work for more insights on the research in this track. 
Though OFA-OCR achieves high accuracy on multiple text recognition datasets, its costs are larger than the non-Transformer baselines. 
As the model is based on Transformer, the costs for training and inference of $\text{OFA-OCR}_\text{base}$ are equivalent to those of other base-size Transformer. 
However, in practice, it is difficult to deploy such large models. 
Thus in our future work, we will discover how to distill or compress OFA-OCR to a light-weight model with high efficiency. 



\section*{Ethics Statement}
Our method is essentially based on a generation model, and thus the OCR results should be taken as AI-generated contents. 
As the generated results should be aligned with the input, we have not noticed deliberate harmful contents, e.g., hate speech, bias, etc.  
However, the model maintains such ability and it might be triggered. 
We also have to note that due to the application context, the input may contain harmful contents itself. In such scenario, we believe the model should faithfully reflect its input, which is valuable to measures against such contents.
Although we believe that after finetuning on the public datasets the risk of such phenomena is extremely low, we still take it into account. 
In the future research, besides focusing on improving downstream performance, we will study how to increase the controllability on the generation.  

\bibliography{anthology,custom}

\begin{thebibliography}{38}
\expandafter\ifx\csname natexlab\endcsname\relax\def\natexlab#1{#1}\fi

\bibitem[{Atienza(2021)}]{vit_ocr}
Rowel Atienza. 2021.
\newblock \href {https://doi.org/10.1007/978-3-030-86549-8\_21} {Vision
  {Transformer} for fast and efficient scene text recognition}.
\newblock In \emph{{ICDAR} {(1)}}, volume 12821 of \emph{Lecture Notes in
  Computer Science}, pages 319--334. Springer.

\bibitem[{Carion et~al.(2020)Carion, Massa, Synnaeve, Usunier, Kirillov, and
  Zagoruyko}]{detr}
Nicolas Carion, Francisco Massa, Gabriel Synnaeve, Nicolas Usunier, Alexander
  Kirillov, and Sergey Zagoruyko. 2020.
\newblock End-to-end object detection with transformers.
\newblock In \emph{European conference on computer vision}, pages 213--229.
  Springer.

\bibitem[{Chen et~al.(2021{\natexlab{a}})Chen, Li, and Xue}]{transocr}
Jingye Chen, Bin Li, and Xiangyang Xue. 2021{\natexlab{a}}.
\newblock \href {https://doi.org/10.1109/CVPR46437.2021.01185} {Scene text
  telescope: Text-focused scene image super-resolution}.
\newblock In \emph{{CVPR}}, pages 12026--12035. Computer Vision Foundation /
  {IEEE}.

\bibitem[{Chen et~al.(2021{\natexlab{b}})Chen, Yu, Ma, Guan, Xu, Wang, Qu, Li,
  and Xue}]{benchmarking_chinese}
Jingye Chen, Haiyang Yu, Jianqi Ma, Mengnan Guan, Xixi Xu, Xiaocong Wang,
  Shaobo Qu, Bin Li, and Xiangyang Xue. 2021{\natexlab{b}}.
\newblock \href {http://arxiv.org/abs/2112.15093} {Benchmarking {Chinese} text
  recognition: Datasets, baselines, and an empirical study}.
\newblock \emph{CoRR}, abs/2112.15093.

\bibitem[{Chen et~al.(2015)Chen, Fang, Lin, Vedantam, Gupta, Doll{\'{a}}r, and
  Zitnick}]{coco_cap}
Xinlei Chen, Hao Fang, Tsung{-}Yi Lin, Ramakrishna Vedantam, Saurabh Gupta,
  Piotr Doll{\'{a}}r, and C.~Lawrence Zitnick. 2015.
\newblock \href {http://arxiv.org/abs/1504.00325} {Microsoft {COCO} captions:
  Data collection and evaluation server}.
\newblock \emph{CoRR}, abs/1504.00325.

\bibitem[{Chen et~al.(2019)Chen, Li, Yu, Kholy, Ahmed, Gan, Cheng, and
  Liu}]{uniter}
Yen-Chun Chen, Linjie Li, Licheng Yu, Ahmed~El Kholy, Faisal Ahmed, Zhe Gan,
  Yu~Cheng, and Jingjing Liu. 2019.
\newblock Uniter: Universal image-text representation learning.
\newblock In \emph{European Conference on Computer Vision}.

\bibitem[{Chng et~al.(2019)Chng, Liu, Sun, Ng, Luo, Ni, Fang, Zhang, Han, Ding,
  Liu, Karatzas, Chan, and Jin}]{art}
Chee-Kheng Chng, Yuliang Liu, Yipeng Sun, Chun~Chet Ng, Canjie Luo, Zihan Ni,
  Chuanming Fang, Shuaitao Zhang, Junyu Han, Errui Ding, Jingtuo Liu,
  Dimosthenis Karatzas, Chee~Seng Chan, and Lianwen Jin. 2019.
\newblock Icdar2019 robust reading challenge on arbitrary-shaped text -
  rrc-art.
\newblock \emph{2019 International Conference on Document Analysis and
  Recognition (ICDAR)}, pages 1571--1576.

\bibitem[{Dosovitskiy et~al.(2021)Dosovitskiy, Beyer, Kolesnikov, Weissenborn,
  Zhai, Unterthiner, Dehghani, Minderer, Heigold, Gelly, Uszkoreit, and
  Houlsby}]{vit}
Alexey Dosovitskiy, Lucas Beyer, Alexander Kolesnikov, Dirk Weissenborn,
  Xiaohua Zhai, Thomas Unterthiner, Mostafa Dehghani, Matthias Minderer, Georg
  Heigold, Sylvain Gelly, Jakob Uszkoreit, and Neil Houlsby. 2021.
\newblock \href {https://openreview.net/forum?id=YicbFdNTTy} {An image is worth
  16x16 words: Transformers for image recognition at scale}.
\newblock In \emph{{ICLR}}. OpenReview.net.

\bibitem[{Fang et~al.(2021)Fang, Xie, Wang, Mao, and Zhang}]{abinet}
Shancheng Fang, Hongtao Xie, Yuxin Wang, Zhendong Mao, and Yongdong Zhang.
  2021.
\newblock Read like humans: Autonomous, bidirectional and iterative language
  modeling for scene text recognition.
\newblock \emph{2021 IEEE/CVF Conference on Computer Vision and Pattern
  Recognition (CVPR)}, pages 7094--7103.

\bibitem[{Goyal et~al.(2017)Goyal, Khot, Summers{-}Stay, Batra, and
  Parikh}]{vqav2}
Yash Goyal, Tejas Khot, Douglas Summers{-}Stay, Dhruv Batra, and Devi Parikh.
  2017.
\newblock \href {https://doi.org/10.1109/CVPR.2017.670} {Making the {V} in
  {VQA} matter: Elevating the role of image understanding in visual question
  answering}.
\newblock In \emph{{CVPR}}, pages 6325--6334. {IEEE} Computer Society.

\bibitem[{Graves et~al.(2006)Graves, Fern{\'a}ndez, Gomez, and
  Schmidhuber}]{ctc}
Alex Graves, Santiago Fern{\'a}ndez, Faustino Gomez, and J{\"u}rgen
  Schmidhuber. 2006.
\newblock Connectionist temporal classification: labelling unsegmented sequence
  data with recurrent neural networks.
\newblock In \emph{Proceedings of the 23rd international conference on Machine
  learning}, pages 369--376.

\bibitem[{He et~al.(2016)He, Zhang, Ren, and Sun}]{resnet}
Kaiming He, Xiangyu Zhang, Shaoqing Ren, and Jian Sun. 2016.
\newblock \href {https://doi.org/10.1109/CVPR.2016.90} {Deep residual learning
  for image recognition}.
\newblock In \emph{{CVPR}}, pages 770--778. {IEEE} Computer Society.

\bibitem[{He et~al.(2018)He, Liu, Yang, Zhang, Luo, Gao, Zheng, Wang, Zhang,
  and Jin}]{mtwi}
Mengchao He, Yuliang Liu, Zhibo Yang, Sheng Zhang, Canjie Luo, Feiyu Gao,
  Qi~Zheng, Yongpan Wang, Xin Zhang, and Lianwen Jin. 2018.
\newblock Icpr2018 contest on robust reading for multi-type web images.
\newblock \emph{2018 24th International Conference on Pattern Recognition
  (ICPR)}, pages 7--12.

\bibitem[{Krishna et~al.(2017)Krishna, Zhu, Groth, Johnson, Hata, Kravitz,
  Chen, Kalantidis, Li, Shamma, Bernstein, and Fei{-}Fei}]{vg}
Ranjay Krishna, Yuke Zhu, Oliver Groth, Justin Johnson, Kenji Hata, Joshua
  Kravitz, Stephanie Chen, Yannis Kalantidis, Li{-}Jia Li, David~A. Shamma,
  Michael~S. Bernstein, and Li~Fei{-}Fei. 2017.
\newblock \href {https://doi.org/10.1007/s11263-016-0981-7} {Visual genome:
  Connecting language and vision using crowdsourced dense image annotations}.
\newblock \emph{Int. J. Comput. Vis.}, 123(1):32--73.

\bibitem[{Li et~al.(2019)Li, Wang, Shen, and Zhang}]{sar}
Hui Li, Peng Wang, Chunhua Shen, and Guyu Zhang. 2019.
\newblock \href {https://doi.org/10.1609/aaai.v33i01.33018610} {Show, attend
  and read: A simple and strong baseline for irregular text recognition}.
\newblock In \emph{{AAAI}}, pages 8610--8617. {AAAI} Press.

\bibitem[{Li et~al.(2021)Li, Lv, Cui, Lu, Flor{\^{e}}ncio, Zhang, Li, and
  Wei}]{trocr}
Minghao Li, Tengchao Lv, Lei Cui, Yijuan Lu, Dinei A.~F. Flor{\^{e}}ncio, Cha
  Zhang, Zhoujun Li, and Furu Wei. 2021.
\newblock \href {http://arxiv.org/abs/2109.10282} {{TrOCR}: Transformer-based
  optical character recognition with pre-trained models}.
\newblock \emph{CoRR}, abs/2109.10282.

\bibitem[{Lin et~al.(2021)Lin, Men, Yang, Zhou, Ding, Zhang, Wang, Wang, Jiang,
  Jia, Zhang, Zhang, Zou, Li, Deng, Liu, Xue, Zhou, Ma, Yu, Li, Lin, Zhou,
  Tang, and Yang}]{m6}
Junyang Lin, Rui Men, An~Yang, Chang Zhou, Ming Ding, Yichang Zhang, Peng Wang,
  Ang Wang, Le~Jiang, Xianyan Jia, Jie Zhang, Jianwei Zhang, Xu~Zou, Zhikang
  Li, Xiaodong Deng, Jie Liu, Jinbao Xue, Huiling Zhou, Jianxin Ma, Jin Yu,
  Yong Li, Wei Lin, Jingren Zhou, Jie Tang, and Hongxia Yang. 2021.
\newblock \href {http://arxiv.org/abs/2103.00823} {{M6}: A {Chinese} multimodal
  pretrainer}.
\newblock \emph{CoRR}, abs/2103.00823.

\bibitem[{Liu et~al.(2019)Liu, Zhang, Zhou, Jiang, Song, Li, Zhou, Wang, Wang,
  Liao, Yang, Bai, Shi, Karatzas, Lu, and Jawahar}]{rects}
Xi~Liu, Rui Zhang, Yongsheng Zhou, Qianyi Jiang, Qi~Song, Nan Li, Kai Zhou, Lei
  Wang, Dong Wang, Minghui Liao, Mingkun Yang, Xiang Bai, Baoguang Shi,
  Dimosthenis Karatzas, Shijian Lu, and C.~V. Jawahar. 2019.
\newblock Icdar 2019 robust reading challenge on reading chinese text on
  signboard.
\newblock \emph{2019 International Conference on Document Analysis and
  Recognition (ICDAR)}, pages 1577--1581.

\bibitem[{Loshchilov and Hutter(2019)}]{adamw}
Ilya Loshchilov and Frank Hutter. 2019.
\newblock Decoupled weight decay regularization.
\newblock In \emph{{ICLR} 2019}.

\bibitem[{Lu et~al.(2019)Lu, Batra, Parikh, and Lee}]{vilbert}
Jiasen Lu, Dhruv Batra, Devi Parikh, and Stefan Lee. 2019.
\newblock Vilbert: Pretraining task-agnostic visiolinguistic representations
  for vision-and-language tasks.
\newblock In \emph{Neural Information Processing Systems}.

\bibitem[{Lu et~al.(2022)Lu, Clark, Zellers, Mottaghi, and
  Kembhavi}]{unified-io}
Jiasen Lu, Christopher Clark, Rowan Zellers, Roozbeh Mottaghi, and Aniruddha
  Kembhavi. 2022.
\newblock Unified-io: A unified model for vision, language, and multi-modal
  tasks.
\newblock \emph{arXiv preprint arXiv:2206.08916}.

\bibitem[{Luo et~al.(2019)Luo, Jin, and Sun}]{moran}
Canjie Luo, Lianwen Jin, and Zenghui Sun. 2019.
\newblock \href {https://doi.org/10.1016/j.patcog.2019.01.020} {{MORAN}: A
  multi-object rectified attention network for scene text recognition}.
\newblock \emph{Pattern Recognit.}, 90:109--118.

\bibitem[{Lyu et~al.(2022)Lyu, Zhang, Liu, Qiao, Xu, Wu, Yao, Han, Ding, and
  Wang}]{maskocr}
Pengyuan Lyu, Chengquan Zhang, Shanshan Liu, Meina Qiao, Yangliu Xu, Liang Wu,
  Kun Yao, Junyu Han, Errui Ding, and Jingdong Wang. 2022.
\newblock \href {https://doi.org/10.48550/arXiv.2206.00311} {{MaskOCR}: Text
  recognition with masked encoder-decoder pretraining}.
\newblock \emph{CoRR}, abs/2206.00311.

\bibitem[{Radford et~al.(2021)Radford, Kim, Hallacy, Ramesh, Goh, Agarwal,
  Sastry, Askell, Mishkin, Clark, Krueger, and Sutskever}]{clip}
Alec Radford, Jong~Wook Kim, Chris Hallacy, Aditya Ramesh, Gabriel Goh,
  Sandhini Agarwal, Girish Sastry, Amanda Askell, Pamela Mishkin, Jack Clark,
  Gretchen Krueger, and Ilya Sutskever. 2021.
\newblock \href {http://proceedings.mlr.press/v139/radford21a.html} {Learning
  transferable visual models from natural language supervision}.
\newblock In \emph{{ICML}}, volume 139 of \emph{Proceedings of Machine Learning
  Research}, pages 8748--8763. {PMLR}.

\bibitem[{Schuhmann et~al.(2022)Schuhmann, Beaumont, Vencu, Gordon, Wightman,
  Cherti, Coombes, Katta, Mullis, Wortsman, Schramowski, Kundurthy, Crowson,
  Schmidt, Kaczmarczyk, and Jitsev}]{laion5b}
Christoph Schuhmann, Romain Beaumont, Richard Vencu, Cade Gordon, Ross
  Wightman, Mehdi Cherti, Theo Coombes, Aarush Katta, Clayton Mullis, Mitchell
  Wortsman, Patrick Schramowski, Srivatsa Kundurthy, Katherine Crowson, Ludwig
  Schmidt, Robert Kaczmarczyk, and Jenia Jitsev. 2022.
\newblock Laion-5b: An open large-scale dataset for training next generation
  image-text models.
\newblock \emph{ArXiv}, abs/2210.08402.

\bibitem[{Shi et~al.(2017{\natexlab{a}})Shi, Bai, and Yao}]{crnn}
Baoguang Shi, Xiang Bai, and Cong Yao. 2017{\natexlab{a}}.
\newblock \href {https://doi.org/10.1109/TPAMI.2016.2646371} {An end-to-end
  trainable neural network for image-based sequence recognition and its
  application to scene text recognition}.
\newblock \emph{{IEEE} Trans. Pattern Anal. Mach. Intell.}, 39(11):2298--2304.

\bibitem[{Shi et~al.(2019)Shi, Yang, Wang, Lyu, Yao, and Bai}]{aster}
Baoguang Shi, Mingkun Yang, Xinggang Wang, Pengyuan Lyu, Cong Yao, and Xiang
  Bai. 2019.
\newblock \href {https://doi.org/10.1109/TPAMI.2018.2848939} {{ASTER}: An
  attentional scene text recognizer with flexible rectification}.
\newblock \emph{{IEEE} Trans. Pattern Anal. Mach. Intell.}, 41(9):2035--2048.

\bibitem[{Shi et~al.(2017{\natexlab{b}})Shi, Yao, Liao, Yang, Xu, Cui,
  Belongie, Lu, and Bai}]{rctw}
Baoguang Shi, Cong Yao, Minghui Liao, Mingkun Yang, Pei Xu, Linyan Cui, Serge
  Belongie, Shijian Lu, and Xiang Bai. 2017{\natexlab{b}}.
\newblock Icdar2017 competition on reading chinese text in the wild (rctw-17).
\newblock In \emph{2017 14th iapr international conference on document analysis
  and recognition (ICDAR)}, volume~1, pages 1429--1434. IEEE.

\bibitem[{Sun et~al.(2019)Sun, Ni, Chng, Liu, Luo, Ng, Han, Ding, Liu,
  Karatzas, Chan, and Jin}]{lsvt}
Yipeng Sun, Zihan Ni, Chee-Kheng Chng, Yuliang Liu, Canjie Luo, Chun~Chet Ng,
  Junyu Han, Errui Ding, Jingtuo Liu, Dimosthenis Karatzas, Chee~Seng Chan, and
  Lianwen Jin. 2019.
\newblock Icdar 2019 competition on large-scale street view text with partial
  labeling - rrc-lsvt.
\newblock \emph{2019 International Conference on Document Analysis and
  Recognition (ICDAR)}, pages 1557--1562.

\bibitem[{Vaswani et~al.(2017)Vaswani, Shazeer, Parmar, Uszkoreit, Jones,
  Gomez, Kaiser, and Polosukhin}]{transformer}
Ashish Vaswani, Noam Shazeer, Niki Parmar, Jakob Uszkoreit, Llion Jones,
  Aidan~N. Gomez, Lukasz Kaiser, and Illia Polosukhin. 2017.
\newblock \href
  {https://proceedings.neurips.cc/paper/2017/hash/3f5ee243547dee91fbd053c1c4a845aa-Abstract.html}
  {Attention is all you need}.
\newblock In \emph{{NIPS}}, pages 5998--6008.

\bibitem[{Wang et~al.(2022{\natexlab{a}})Wang, Yang, Men, Lin, Bai, Li, Ma,
  Zhou, Zhou, and Yang}]{ofa}
Peng Wang, An~Yang, Rui Men, Junyang Lin, Shuai Bai, Zhikang Li, Jianxin Ma,
  Chang Zhou, Jingren Zhou, and Hongxia Yang. 2022{\natexlab{a}}.
\newblock \href {https://proceedings.mlr.press/v162/wang22al.html} {{OFA}:
  Unifying architectures, tasks, and modalities through a simple
  sequence-to-sequence learning framework}.
\newblock In \emph{{ICML}}, volume 162 of \emph{Proceedings of Machine Learning
  Research}, pages 23318--23340. {PMLR}.

\bibitem[{Wang et~al.(2022{\natexlab{b}})Wang, Bao, Dong, Bjorck, Peng, Liu,
  Aggarwal, Mohammed, Singhal, Som, and Wei}]{beit3}
Wenhui Wang, Hangbo Bao, Li~Dong, Johan Bjorck, Zhiliang Peng, Qiang Liu, Kriti
  Aggarwal, Owais~Khan Mohammed, Saksham Singhal, Subhojit Som, and Furu Wei.
  2022{\natexlab{b}}.
\newblock \href {https://doi.org/10.48550/arXiv.2208.10442} {Image as a foreign
  language: {BEiT} pretraining for all vision and vision-language tasks}.
\newblock \emph{CoRR}, abs/2208.10442.

\bibitem[{Wang et~al.(2021)Wang, Yu, Yu, Dai, Tsvetkov, and Cao}]{simvlm}
Zirui Wang, Jiahui Yu, Adams~Wei Yu, Zihang Dai, Yulia Tsvetkov, and Yuan Cao.
  2021.
\newblock Simvlm: Simple visual language model pretraining with weak
  supervision.
\newblock \emph{ArXiv}, abs/2108.10904.

\bibitem[{Yu et~al.(2020)Yu, Li, Zhang, Liu, Han, Liu, and
  Ding}]{semantic_reasoning}
Deli Yu, Xuan Li, Chengquan Zhang, Tao Liu, Junyu Han, Jingtuo Liu, and Errui
  Ding. 2020.
\newblock \href {https://doi.org/10.1109/CVPR42600.2020.01213} {Towards
  accurate scene text recognition with semantic reasoning networks}.
\newblock In \emph{{CVPR}}, pages 12110--12119. Computer Vision Foundation /
  {IEEE}.

\bibitem[{Yu et~al.(2016)Yu, Poirson, Yang, Berg, and Berg}]{refcoco}
Licheng Yu, Patrick Poirson, Shan Yang, Alexander~C. Berg, and Tamara~L. Berg.
  2016.
\newblock \href {https://doi.org/10.1007/978-3-319-46475-6\_5} {Modeling
  context in referring expressions}.
\newblock In \emph{{ECCV} {(2)}}, volume 9906 of \emph{Lecture Notes in
  Computer Science}, pages 69--85. Springer.

\bibitem[{Yuan et~al.(2019)Yuan, Zhu, Xu, Li, Mu, and Hu}]{ctw}
Tailing Yuan, Zhe Zhu, Kun Xu, Cheng-Jun Li, Tai-Jiang Mu, and Shimin Hu. 2019.
\newblock A large chinese text dataset in the wild.
\newblock \emph{Journal of Computer Science and Technology}, 34:509--521.

\bibitem[{Zhang et~al.(2020)Zhang, Liang, and Jin}]{hccdoc}
Hesuo Zhang, Lingyu Liang, and Lianwen Jin. 2020.
\newblock Scut-hccdoc: A new benchmark dataset of handwritten chinese text in
  unconstrained camera-captured documents.
\newblock \emph{Pattern Recognit.}, 108:107559.

\bibitem[{Zhang et~al.(2022)Zhang, Zhu, Yao, Sun, Li, and Yu}]{conclr}
Xinyun Zhang, Binwu Zhu, Xufeng Yao, Qi~Sun, Ruiyu Li, and Bei Yu. 2022.
\newblock \href {https://ojs.aaai.org/index.php/AAAI/article/view/20245}
  {Context-based contrastive learning for scene text recognition}.
\newblock In \emph{{AAAI}}, pages 3353--3361. {AAAI} Press.

\end{thebibliography}
\bibliographystyle{acl_natbib}

\newpage
\appendix

\section{Appendix}
\label{sec:appendix}

\subsection{Datasets}
\label{datasets}
The Chinese text recognition benchmark consists of $4$ subtasks, which are scene, web, document, and handwriting. 
The scene dataset consists of multiple datasets, including RCTW~\citep{rctw}, ReCTS~\citep{rects}, LSVT~\citep{lsvt}, ArT~\citep{art}, and CTW~\citep{ctw}. 
It consists of \num{509164} samples for training, \num{63645} for validation, and \num{63646} for testing. 
The web dataset is derived from MTWI~\citep{mtwi}, and it has \num{112471} samples for training, \num{14059} for validation, and \num{14059} for testing. 
The document dataset is constructed with synthetic data created with Text Renderer\footnote{\url{https://github.com/Sanster/text_renderer}}, and it has \num{400000} samples for training, \num{50000} for validation, and \num{50000} for testing. 
The handwriting dataset is collected from SCUT-HCCDoc~\citep{hccdoc}, and it has \num{74603} samples for training, \num{18651} for validation, and \num{23389} for testing. 

\subsection{Evaluation}
We calculate the ratio of exact match as the accuracy for the evaluation. 
For the average score on the $4$ subtasks, we calculate the average score weighted by the number of testing samples~\citep{maskocr}.

\subsection{Implementation Details}
\label{implementation_details}
For single-task, multitask, and multitask + single-task finetuning, we finetune the pretrained base-size and large-size OFA for \num{100} epochs. 
We use the AdamW~\citep{adamw} optimizer for training. 
For the base-size model, the batch size is \num{256} and the peak learning rate is \num{5e-5}, and for the large-size model, the batch size is \num{512} and the peak learning rate is \num{2e-5}. 

Here we provide more details about the preprocessing for images. The specified resolution is $480 \times 480$, and as the pretrained models were pretrained on images of the resolution of $224 \times 224$, we apply interpolation to the positional embedding. 

As to the data augmentation, we demonstrate the process with the pseudo code below. 

\begin{lstlisting}
import torch
from torchvision.transforms import InterpolationMode
from torchvision.transforms import functional as F

def ocr_resize(img, resolution=480, is_document=False):
    img = img.convert("RGB")
    width, height = img.size

    if width >= height:
        new_width = max(64, resolution)
        new_height = max(64, int(resolution * (height / width)))
        top = random.randint(0, resolution - new_height)
        bottom = resolution - new_height - top
        left, right = 0, 0
    else:
        new_height = max(64, resolution)
        new_width = max(64, int(resolution * (width / height)))
        left = random.randint(0, resolution - new_width)
        right = resolution - new_width - left
        top, bottom = 0, 0

    img_new = F.resize(
        img,
        [new_height, new_width],
        interpolation=InterpolationMode.BICUBIC,
    )

    img_new = F.pad(img_new, padding=[left, top, right, bottom], padding_mode="edge")

    return img_new
\end{lstlisting}

\subsection{Case Study}
\label{case_study}

\begin{figure*}[t] 
    \centering
    \includegraphics[width=1.0\linewidth]{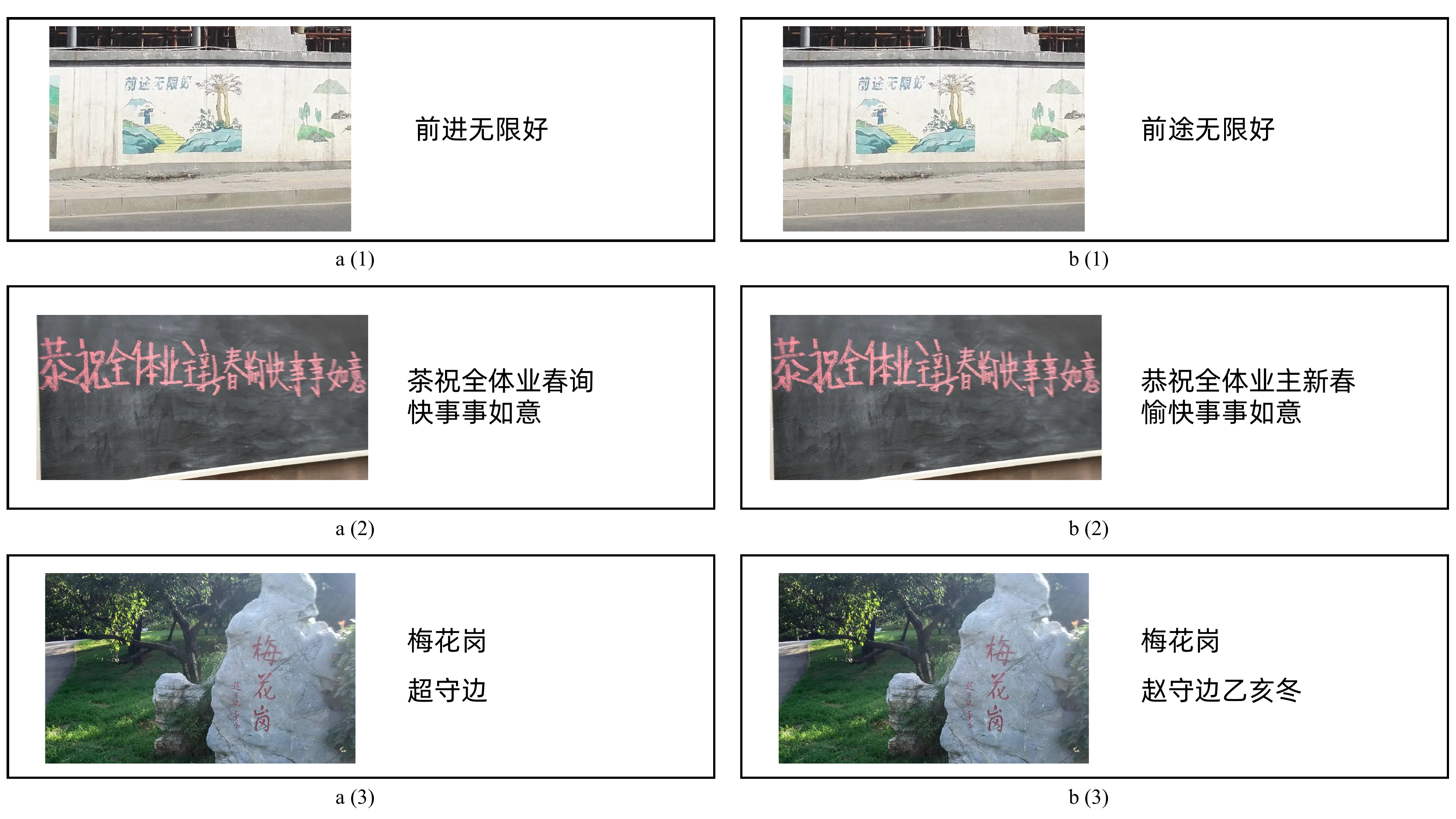}
    \caption{\textbf{A Case study of different OCR demos.} We compare a product-level API (a) with OFA-OCR (b). Through the case study, we find that OFA-OCR can reach a competitive performance. }
    \label{fig:case}
\end{figure*}

Here we evaluate the performance of the constructed simple OCR pipeline. For comparison, we use a product-level API\footnote{\url{https://www.paddlepaddle.org.cn/modelsDetail?modelId=17}} as the baseline. Figure~\ref{fig:case} demonstrates the cases comparison. It can be found that on the $3$ cases while the baseline makes mistakes by different extents, OFA-OCR makes the correct prediction of all characters, even if there are missing strokes or the text is in hard-to-recognize handwriting style. 

\end{document}